\documentclass[11pt]{article}
\usepackage{coling2020}
\usepackage{times}
\usepackage{mathptmx}
\usepackage{url}
\usepackage{latexsym}

\usepackage{subcaption}
\usepackage{graphicx}
\usepackage{placeins}
\usepackage{enumitem}

\DeclareCaptionFont{tiny}{\tiny}

\usepackage{amsmath,amsthm,amssymb,amsfonts,mathtools}

\usepackage{tabularx}
\usepackage{multirow}
\usepackage{wrapfig}

\usepackage[justification=justified,singlelinecheck=false]{caption}

\usepackage{algorithm}% http://ctan.org/pkg/algorithms
\usepackage{algpseudocode}% http://ctan.org/pkg/algorithmicx

\usepackage{tikz-dependency}
\usepackage{tikz}
\usetikzlibrary{positioning}
\definecolor {processblue}{cmyk}{0.96,0,0,0}

\colingfinalcopy % Uncomment this line for the final submission

\title{Aspectuality Across Genre: A Distributional Semantics Approach}

\author{Thomas Kober \\
  Rasa \\
  {\tt t.kober@rasa.com} \\\And
  Malihe Alikhani \\
  University of Pittsburgh \\
  {\tt malihe@pitt.edu} \\\AND
  Matthew Stone \\
  Rutgers University \\
  {\tt mdstone@rutgers.edu} \\\And
  Mark Steedman \\
  University of Edinburgh \\
  {\tt steedman@inf.ed.ac.uk}
\\}
\date{}

\begin{document}
\maketitle
\begin{abstract}
The interpretation of the lexical aspect of verbs in English plays a crucial role for recognizing textual entailment and learning discourse-level inferences.
We show that two elementary dimensions of aspectual class, \emph{states} vs. \emph{events}, and \emph{telic} vs. \emph{atelic} events, can be modelled effectively with distributional semantics. We find that a verb's local context is most indicative of its aspectual class, and demonstrate that closed class words tend to be stronger discriminating contexts than content words. Our approach outperforms previous work on three datasets. Lastly, we contribute a dataset of human--human conversations annotated with lexical aspect and present experiments that show the correlation of telicity with genre and discourse goals. 
\end{abstract}

\section{Introduction}

% \begin{itemize}
%     \item why? discourse, dialogue, semantics, entailment and more
%     \item what we do? genre, state vs event, telic vs atelic, dataset, identifying challenges moving forward
%     \item entailment graphs
%     \item multimodal analysis and discourse: the role of telicity 
%     \item translation: the role of telicity
%     \item add an example in the intro for motivation
% \end{itemize}

One of the fascinating aspects of studying aspectual class of verbs in English is its relation with non-verbal categories. Thus, although in origin a property of the verb, the aspectual class interacts in a tight-knit fashion with other words in a sentence. Previous research has discussed the importance of predicting the aspectual classes of verbs for predicting coherence relations in text and imagery~\cite{alikhani-stone-2019-caption}, predicting links in entailment graphs~\cite{hosseini2019duality} and interpreting sign languages~\cite{wilbur2003representations}.
In addition, knowledge about the aspectual class of a verb phrase, and its influence on the temporal extent and entailments that it licenses, has been leveraged in the past for a number of natural language understanding tasks such as temporal relation extraction~\cite{Costa_2012}, event ordering~\cite{Chambers_2014,Modi_2014}, and statistical machine translation~\cite{Loaiciga_2016}. %Figure~\ref{} THOMAS PLEASE ADD HERE 
%shows an example of 

% Determining key interpretations of verbs in context, entailment graphs, link prediction models and more generally natural language understating models are incomplete without an accurate model that determines the lexical aspect of verbs in context.

The \emph{Aktionsart}~\cite{Vendler_1957} of a verb
% of a verb phrase is one of the most important factors to interpret the meaning of a sentence. It 
determines the temporal extent of the predication as well as whether it causes a change of state for the entities involved~\cite{Filip_2012}. As \emph{Aktionsart} typically refers to the lexical aspect of a verb in isolation, we adopt the terminology of Verkuyl~\shortcite{Verkuyl_2005}, and refer to the compositionally formed \emph{Aktionsart} of a verb phrase as \emph{predicational aspect}.

One of the most important distinctions of the predicational aspect of a verb is between \emph{\textbf{states}}, such as \emph{to know} or \emph{to love}, and \emph{\textbf{events}}, such as \emph{visit} or \emph{swim}. This distinction is important for identifying the entailments that a given verb phrase licenses, as \emph{stative} predications do not, by definition, entail any change of state. This property has important consequences for a number of natural language understanding tasks such as question answering. For example, if it is known that \emph{John has arrived in Vienna}, a system leveraging aspectual information will be able to infer that the completion of the event of \emph{arriving in Vienna}, indicated by the perfect VP \emph{having arrived in}, has caused a change of state which entails \emph{being in}. Therefore, when asked \emph{Where is John?}, the system will be able to produce the correct answer: \emph{Vienna}. On the other hand, a predominantly stative verb such as \emph{to know}, as in \emph{Eve knows a lot about quantum mechanics}, does not cause a change of state for either \emph{Eve} or \emph{quantum mechanics}. 

Telic predicates do not license consequent state inferences from their progressive VP forms to corresponding non-progressive forms.\footnote{This is also known as the \emph{Imperfective Paradox}~\cite{Dowty_1979}.} Thus, telic/atelic classifications are supported by contrastive pairs like the following:

\begin{enumerate}[label={(\arabic*)}]
    \item Mary \emph{was drawing} a circle $\nrightarrow$ Mary \emph{drew} a circle (telic)
    \item Mary \emph{was pushing} a cart $\rightarrow$ Mary pushed a cart (atelic)
\end{enumerate}
%\begin{itemize}
%    \item Mary was drawing a circle $\nrightarrow$  Mary drew a circle (telic)
%    \item Mary was pushing a cart $\rightarrow$ Mary pushed a cart (atelic)
%\end{itemize}

% MOTIVATE FURTHER FOR APPLICATIONS WHERE THIS IS IMPORTANT, EXAMPLE LIKE QUESTION ANSWERING? MODELLING DIALOGUE? KEEPING A KNOWLEDGE BASE UP TO DATE?

In this paper we propose to approach the problem of classifying predicational aspect with distributional semantics. Our hypothesis is that the meaning distinctions of a verb that relate to its aspectual class should be reflected in its distribution when composed with its context. We therefore intersect word vectors with their context in order to determine a VP's predicational aspect, and show that we achieve a new state-of-the-art on two datasets. We further evaluate our approach on two new genres: image captions and situated human--human conversations, thereby extending the validity of our findings across a variety of genres.

% WE FURTHERMORE SHOW THAT THIS KNOWLEDGE IS USEFUL FOR NATURAL LANGUAGE INFERENCE, PROPOSE NEW DATASET, IMPROVE MODELS

% We first ANALYSE WHICH CONTEXTS OF A VERB ARE MOST IMPORTANT FOR DETERMINING ITS ASPECTUAL CLASS AND ACHIEVE SOTA ON 2 DATASETS WITH A LOCAL COMPOSITIONAL APPROACH. SUBSEQUENTLY WE INTRODUCE A NEW ENTAILMENT DATASET BASED ON AKTIONSART AND DEMONSTRATE HOW SOTA PRETRAINED MODELS FAIL TO CAPTURE INFERENCES THAT REQUIRE REASONING WITH AKTIONSART ON OUR NEW DATASET AND AN EXISTING DATASET. FINALLY WE SHOW HOW OUR PRE-TRAINED ASPECT CLASSIFIER CAN BE TRANSFERRED/COMBINED WITH THE OTHER MODELS IN ORDER TO SUBSTANTIALLY IMPROVE PERFORMANCE ON INFERENCES REQUIRING AKTIONSART REASONING.

% WHILE IT HAS FREQUENTLY BEEN ARGUED THAT THE PERFECT INTRODUCES A STATE TO A DISCOURSE (AND THEREFORE SHOULD BE TREATED AS SUCH), WE FOLLOW MOENS AND STEEDMAN AND ARGUE THAT THE PERFECT INTRODUCES A CONSEQUENT STATE, CAUSED BY THE EVENT DESCRIBED IN THE PERFECT. THIS ALSO AGREES WITH CARLOTA SMITH THAT TREATS THE NUCLEUS OF THE ACTION IN THE PERFECT AS EVENTIVE. FOR EXAMPLE, \emph{has bought} in \emph{Jane has bought the house} WILL BE TREATED AS THE CAUSE OF THE CONSEQUENT STATE (\emph{Jane owns the hose}).

% STRUCTURE:
% BACKGROUND AKTIONSART (AND ITS RELEVANCE TO ENTAILMENT)
% MODELLING PREDICATIONAL ASPECT (COVER DATASETS, EXPERIMENTS AND ANALYSIS)
% ENTAILMENT (INTRODUCING DATASETS, EXPERIMENTS AND ANALYSIS)
% RELATED WORK AT THE END

\section{Related Work}
\label{related}

An early approach to classifying the lexical aspectual class of a verb in context was proposed by Passonneau~\shortcite{Passonneau_1988}, who applied a decompositional analysis of the verb to determine the aspectual class for verb occurrences in a restricted domain. 
The first general-purpose study was conducted by Siegel and McKeown~\shortcite{Siegel_2000}, who built up on earlier work by Klavans and Chodorov~\shortcite{Klavans_1992}, and collected \emph{linguistic indicators} for lexical aspect from a large corpus. These include the presence of \emph{in}- or \emph{for}-adverbials, the tense of the verb or its frequency. Siegel and McKeown~\shortcite{Siegel_2000} subsequently applied different supervised machine learning algorithms to classify the extracted feature vectors into either \emph{states} or \emph{events}, or \emph{telic} or \emph{atelic} events. Siegel and McKeown~\shortcite{Siegel_2000} show that their method substantially improves over a majority-class baseline. The first approach to include features derived from a distributional semantic model has been proposed by Friedrich and Palmer~\shortcite{Friedrich_2014}. In addition to the linguistic indicator features of Siegel and McKeown~\shortcite{Siegel_2000}, Friedrich and Palmer~\shortcite{Friedrich_2014} extract representative stative, dynamic or mixed verbs from the \emph{lexical conceptual structure (LCS)} database~\cite{Dorr_1997} and subsequently use distributional representations to derive similarity scores for the mined verbs.

Another extension to the work of Friedrich and Palmer~\shortcite{Friedrich_2014} has been proposed by Heuschkel~\shortcite{Heuschkel_2016}, who refines the distributional similarity features by first contextualising a target verb with its subject or object, and only then computing the distributional similarities to the set of representative verbs from the LCS database as in Friedrich and Palmer~\shortcite{Friedrich_2014}. All else being equal, Heuschkel~\shortcite{Heuschkel_2016} shows that contextualising the distributional representations improves performance on the \textbf{Asp-ambig} dataset of Friedrich and Palmer~\shortcite{Friedrich_2014}.

In contrast to this line of research we do not make explicit use of any hand-engineered linguistic indicator features but show that these can be picked up in an unsupervised way by composing distributional semantic word representations. The linguistic indicators are furthermore frequently collected on the verb \emph{type} level instead of on the \emph{token} level. Similar to Falk and Martin~\shortcite{Falk_2016}, we are concerned with classifying \emph{verb readings}; however, we do not use engineered features as Falk and Martin~\shortcite{Falk_2016} do, but directly leverage local contextual information in the form of distributional representations. Our approach is also not reliant on the availability of a parallel corpus as in~\newcite{Friedrich_2017}.  The major difference between our approach of using distributional word representations and previous approaches is that we are using the word representations \emph{directly} for classification, rather than \emph{indirectly} by computing similarity scores and using these as features. This furthermore liberates us from the requirement of having a representative seed set of verbs per class to compute the distributional similarities from.
\section{Dataset}
\label{dataset}
We introduce a new dataset, \textbf{DIASPORA},\footnote{\textbf{DI}alogue \textbf{ASP}ectuality from \textbf{ORA}l conversations. The dataset is publicly available from \url{https://go.rutgers.edu/cb6le5c1}.} of human-human conversations annotated with predicational aspect, representing the first dialogue dataset annotated with aspectual information. As reviewed in Section~\ref{related}, what partly makes computational research on predicational aspect difficult is the lack of diverse annotated corpora. With the release of \textbf{DIASPORA} we provide a situated conversational perspective on predicational aspect research, and thereby extend the existing evaluation repertoire by a very important genre.

\paragraph{Annotation effort.}

As our starting point, we sampled 2000 utterances from the Walking Around corpus~\cite{Brennan_2013} uniformly at random. The Walking Around corpus is a dataset of human--human phone conversations where one party needs to find certain landmarks on a university campus and receives directions via phone from the other party. Table~\ref{tbl:walking_around_example} lists part of a conversation from the Walking Around corpus, where the speakers identify a landmark that the second speaker needs to reach. 
\begin{table*}[!htb]
\centering
\small
%\resizebox{\textwidth}{!}{
\begin{tabular}{ l l}
...\\
Speaker 1 & \emph{You're looking for the ship scul- sculpture} \\
Speaker 2 & \emph{Okay} \\
Speaker 1 & \emph{it should be (..) yeah} \\
Speaker 2 & \emph{I think I see it but I'm not close enough yet} \\
Speaker 1 & \emph{It should be right in front of the Heavy Engineering Building} \\
Speaker 2 & \emph{It's a ship?} \\
Speaker 1 & \emph{Yeah it looks like one (..) like it's got (..) how do you describe i- it looks really weird} \\
...
\end{tabular}%}
\caption{Part of an example dialogue from the Walking Around corpus~\cite{Brennan_2013}.}
\label{tbl:walking_around_example}
\end{table*}

We chose the Walking Around corpus because its conversations are situated and in real-time, and because it contains a good distribution of \emph{stative}, \emph{telic}, and \emph{atelic} verb phrases. After sampling the initial set of 2000 utterances, we filtered multi-sentence utterances and utterances that did not contain a verb. We furthermore removed any filled pauses (indicated by ``\emph{(..)}" in Table~\ref{tbl:walking_around_example}) that have been transcribed and marked in the dataset. Following Alikhani and Stone~\shortcite{alikhani-stone-2019-caption}, we annotated the \emph{first} VP for predicational aspect in all utterances. For example, the last utterance of Speaker 1 in Table~\ref{tbl:walking_around_example} contains multiple verbs, and we have annotated the phrase \emph{Yeah it looks like one}---the first VP in the utterance.

The study has been approved by Rutgers’s IRB. Expert annotators annotated the whole dataset and were paid an hourly rate of 15 USD. They were final year linguistics undergraduate students and were provided with an annotation protocol for their task.\footnote{The annotation protocol is published with the dataset at \url{https://go.rutgers.edu/cb6le5c1}.}  To assess the inter-annotator agreement, we determine Cohen’s $\kappa$ value. We randomly selected 200 sentences and assigned each to two annotators, obtaining a Cohen’s $\kappa$ of 0.81, which indicates almost perfect agreement~\cite{Viera_2005}.

\paragraph{Overall statistics.}

The final \textbf{DIASPORA} dataset contains 927 annotated utterances, consisting of 400 utterances labelled as expressing \emph{stative} predicational aspect (43\%), 279 labelled as \emph{telic} (30\%), and 248 labelled as \emph{atelic} (27\%). The overall average utterance length is 15.58. 
\begin{table*}[!htb]
\centering
\small
%\resizebox{\textwidth}{!}{
\begin{tabular}{ l | c | c | c | c}
\textbf{Label} & \textbf{Mean} & \textbf{Median} & \textbf{Min} & \textbf{Max} \\\hline
State & 16.34 & 13 & 4 & 94 \\
Telic & 14.65 & 11 & 3 & 80 \\
Atelic & 15.38 & 12 & 2 & 74 \\\hline
\end{tabular}%}
\caption{Utterance length statistics per label in \textbf{DIASPORA}.}
\label{tbl:diaspora_utterance_len_stats}
\end{table*}
Table~\ref{tbl:diaspora_utterance_len_stats} lists length statistics for \textbf{DIASPORA} per individual label. The means and medians are relatively similar across all classes, suggesting that there is no bias in terms of utterance lengths for any individual class.

The \textbf{DIASPORA} dataset contains 98 unique verb forms, spanning 69 lemmas, with the top 10 most frequent verb lemmas making up $\approx$78\% of all verbs in the corpus. 
\begin{figure*}[!htb]
\centering
\includegraphics[width=\textwidth]{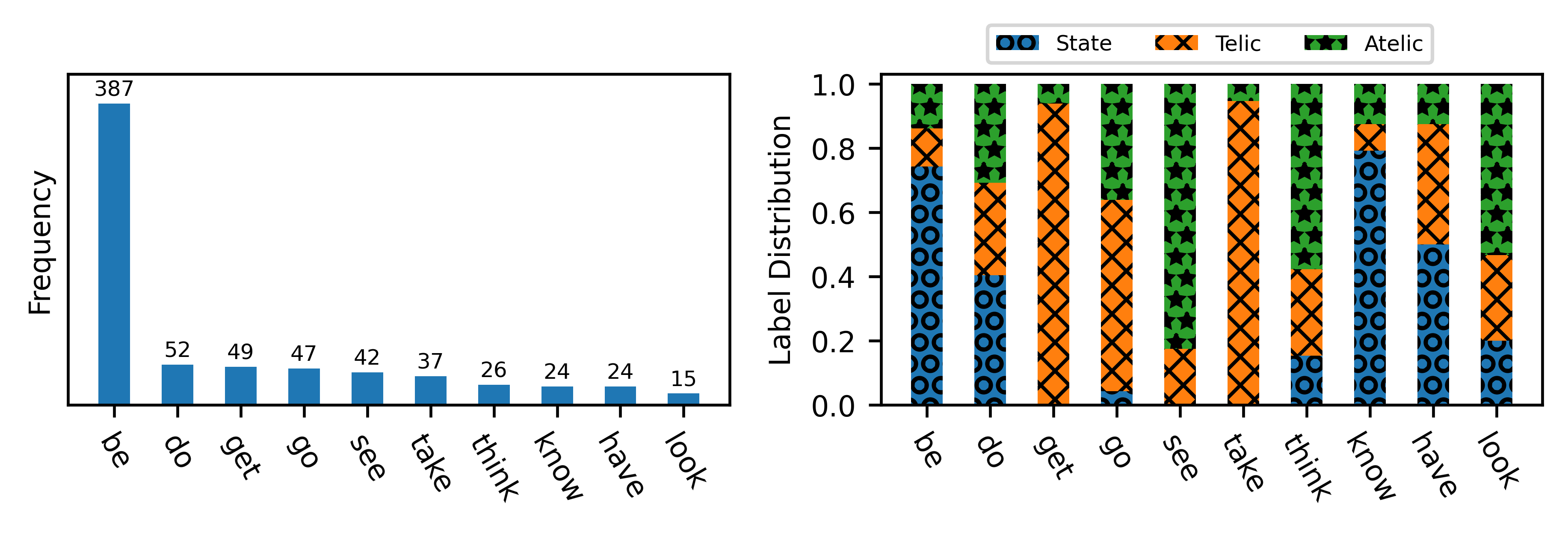}
\captionsetup{font=small}
\caption{Frequency distribution of the 10 most frequent verbs (left) and associated label distribution for the 10 most frequent verbs (right).}
\label{fig:diaspora_verb_and_label_dist}
\end{figure*}

The characteristic of few verbs making up a large proportion of the overall data has already been observed for captions~\cite{alikhani-stone-2019-caption}. This is an expected property in \textbf{DIASPORA} and is due to the single-domain nature of the Walking Around corpus. Figure~\ref{fig:diaspora_verb_and_label_dist} shows the frequency distribution of the top 10 most frequent verbs and their associated label distribution. The large proportion of various forms of \emph{be} is due to many utterances of either speaker referring to the current location of the subject looking for the landmark (e.g. utterances like \emph{I am currently at \dots}). The label distribution of the 10 most frequent verbs shows that there are some highly skewed verbs, such as \emph{get}, \emph{see} or \emph{know}, which have a clear majority class, whereas \emph{do} or \emph{look} exhibit a much more balanced, and therefore ambiguous label distribution. 
\section{Experiments}
%The utility of distributional semantic word representations --- both high-dimensional count-based, and low-dimensional neural word embeddings --- has been shown in a large body of works in recent years at the lexical~\cite{Baroni_2014,Weeds_2014b,Roller_2016,Nguyen_2017}, as well as the sentence level~\cite{Socher_2013b,Bowman_2015,Mihaylov_2017}. In order to contextualise a verb within a phrase we apply pointwise addition as a simple distributional composition function. Pointwise addition in neural word embeddings approximates the intersection of their contexts~\cite{Tian_2017}, and has been shown to be an efficient function for contextualising a word with its phrasal context~\cite{Arora_2016,Kober_2017}. 
% TODO: The first sentence needs to be re-formulated
The utility of distributional semantic word representations has been shown in a large body of works in recent years (Weeds et al.~\shortcite{Weeds_2014b}, Nguyen et al.~\shortcite{Nguyen_2017}, Socher et al.~\shortcite{Socher_2013b}, Bowman et al.~\shortcite{Bowman_2015}; \emph{passim}). 
In order to compose a verb with its context we apply pointwise addition as a simple distributional composition function. Pointwise addition in neural word embeddings approximates the intersection of their contexts\footnote{Note that for sparse count-based distributional representations, addition would correspond to a union of contexts and multiplication to an intersection~\cite{Kober_2018}.}~\cite{Tian_2017}, and has been shown to be an efficient function for contextualising a word in a phrase~\cite{Arora_2016,Kober_2017}. 

\subsection{Distributional Models for Predicational Aspect}

Following previous work on modelling the aspectual class of a verb~\cite{Siegel_2000,Friedrich_2014}, we treat the problem as a supervised classification task, $y = f(x)$, where $y$ represents the aspectual class of a verb, $f$ represents a classification algorithm, and $x$ an input vector representation of a verb in context. For all of our experiments, $f$ is a logistic regression classifier,\footnote{We use scikit-learn~\cite{Pedregosa_2011}---which, much like ourselves, is relying on numpy~\cite{Harris_2020}.} with default hyperparameter settings. In all our experiments, the input vector $x$ is based on $300$-dimensional pre-trained skip-gram word2vec~\cite{Mikolov_2013b} vectors.\footnote{The pre-trained vectors are availble from~\url{https://code.google.com/archive/p/word2vec/}, and we used gensim~\cite{Rehurek_2010} for processing the word2vec representations.} We lowercase all words, but do not apply any other form of morphological preprocessing, which means that we retain different representations for different inflected forms of a verb---i.e. \emph{look}, \emph{looks}, \emph{looking}, and \emph{looked} are represented by 4 distinct vectors.

\subsubsection{Classifying Aspect with Distributional Semantics}

For this approach we obtain a word2vec representation $x$ for a given verb $v$ and feed $x$ into a logistic regression classifier in order to predict the aspectual class of $v$. This approach represents a rather na\"{i}ve baseline that assumes that the aspectual class of a verb is a purely lexical phenomenon on the type level and can be determined independent of any context.

\subsubsection{Incorporating Context with Distributional Composition}

Let $x$ be a word2vec representation for a given verb $v$, and $C$ be the set of context words extracted for $v$, with $c \in C$ denoting the vector representation for an extracted context word of $v$. The composed representation of $v$, denoted by $x^{\prime}$, can then be expressed as a simple sum:
\begin{equation}
\label{eqn:composition}
	x^{\prime} = x + \sum_{c \in C}{c}
\end{equation}
Subsequently, $x^{\prime}$ is passed through a logistic regression classifier in order to predict the aspectual class of $v$. This model aims to capture the compositional nature of predicational aspect by integrating local contextual information into the model.

\subsubsection*{Types of Context}

\begin{figure}[!htb]
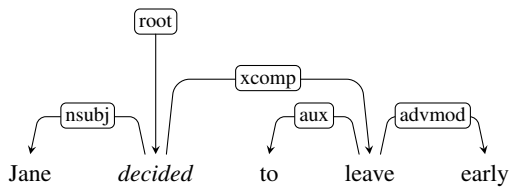

\centering
\begin{dependency}
\begin{deptext}[column sep=0.7cm]
  \small{Jane}\&
    \small{\emph{decided}}\& 
      \small{to}\&
        \small{leave}\&
          \small{early}\\
\end{deptext}
\deproot{2}{root}
\depedge{2}{1}{nsubj}
\depedge{2}{4}{xcomp}
\depedge{4}{3}{aux}
\depedge{4}{5}{advmod}
\end{dependency}
\captionsetup{font=small}
\caption{With a linear context window of size $1$, \emph{Jane} and \emph{to} would be extracted as contexts for the verb \emph{decided}. With a dependency-based context, \emph{Jane} and \emph{leave} would be extracted.}
\label{fig:example_sentence_context}
\end{figure}

We investigate two different kinds of context: simple linear context windows of varying length and first-order dependency contexts. For example for the sentence in Figure~\ref{fig:example_sentence_context}, a linear context window of size 1 would extract \emph{Jane} and \emph{to} for the target verb \emph{decided}, whereas a dependency-based context would extract \emph{Jane} and \emph{leave}. We used the Stanford NLP pipeline~\cite{Manning_2014} with default settings for parsing the sentences in our datasets. 

For linear context windows we use sizes \{1, 2, 3, 5, 10\}, and for first-order dependency-based contexts we experiment with using only the head\footnote{If the target verb is the root, then no context is used for that verb.} of the verb, only its children, or the full first-order context. 

\subsubsection*{Incorporating the Full Sentence}

We furthermore test a model that incorporates the whole sentential context into a vector representation. The approach simply uses all words from a given sentence and composes their corresponding word2vec representations as in Equation~\ref{eqn:composition} above to create an embedding for the whole sentence. Embedding a sentence by adding word vectors has been shown to be an effective method for other NLP tasks such as sentiment analysis~\cite{Iyyer_2015} and recognising textual entailment~\cite{Wieting_2016b}.

The underlying rationale behind this approach is that the aspectual class of a verb is a function of the sentence as a whole, rather than dependent on local context alone~\cite{Moens_1987,Moens_1988,Dowty_1991}.%TIME PERMITTING INFERSENT ENCODER % https://github.com/facebookresearch/InferSent

\section{Experiments}
\label{sec:experiments}

We perform experiments that assess the suitability of distributional representations for distinguishing \emph{states} from \emph{events} (\S~\ref{sec:exp_states_vs_events}), and \emph{telic} from \emph{atelic} events (\S~\ref{sec:exp_telic_vs_atelic}). Only a completed and \emph{telic} event licenses a new consequent \emph{state}. Therefore, modelling predicational aspect is important for deeper text understanding, for example for modelling cause and effect, and especially for inferring consequent states. 

\subsection{Experiment 1 --- States vs. Events}
\label{sec:exp_states_vs_events}

For the distinction between \emph{states} and \emph{events} we perform experiments on 5 datasets in total. We use the \textbf{Asp-ambig} dataset by Friedrich and Palmer~\shortcite{Friedrich_2014}, the \textbf{SitEnt} dataset by Friedrich et al.~\shortcite{Friedrich_2016}, our own sub-sampled version of the \textbf{SitEnt} dataset, the \textbf{Captions} dataset by Alikhani and Stone~\shortcite{alikhani-stone-2019-caption}, and our own \textbf{DIASPORA} dataset, proposed in this work. 

The \textbf{Asp-ambig} dataset is sampled from the Brown corpus~\cite{Francis_1979} and is based on 20 frequently occurring verbs whose predicational aspect changes depending on context. For each verb, Friedrich and Palmer~\shortcite{Friedrich_2014} collected 138 sentences, resulting in 2760 examples in total. The dataset contains the annotations of whether the verb in context expresses a \emph{state}, \emph{event}, or whether it could be \emph{both}.\footnote{The instances labelled as both have either been labelled by the annotators as both because both readings are possible, or have been labelled as both when the annotators disagreed in their judgement~\cite{Friedrich_2014}.} Following Friedrich and Palmer~\shortcite{Friedrich_2014}, we report accuracy using leave-one-out cross-validation.\footnote{A performance overview per verb type is included in Appendix~\ref{sec:supplemental_a}}

We furthermore evaluate our approach on the \textbf{SitEnt} dataset~\cite{Friedrich_2016}. The \textbf{SitEnt} dataset contains ~40k sentences from the MASC corpus~\cite{Ide_2008} and English Wikipedia, split into separate training and test sets. We evaluate our approach on the test set, using the original split of Friedrich et al.~\shortcite{Friedrich_2016}. The \textbf{SitEnt} dataset contains annotations for verbs in context as either expressing a \emph{state} or an \emph{event}, but not both. Following Friedrich et al.~\shortcite{Friedrich_2016}, we report class-based F1-scores.

During model development we noticed an idiosyncrasy in the \textbf{SitEnt} dataset, where only ~900 verb types out of ~4.5k occurred with both class labels, and only 267 of them had a balanced (i.e. ambiguous) class distribution. We considered this as likely problematic as a classifier might just pick up this artefact. We therefore created a downsampled dataset --- \textbf{SitEnt-ambig} --- that only contains verb types with a balanced class distribution, randomly sub-sampling the majority class for verb types with an imbalanced class distribution\footnote{We allowed the imbalance to be at most 60:40.}. The resulting dataset consists of 6547 examples in the training set and 1402 examples in the test set.\footnote{The sub-sampled dataset is available from \url{https://go.rutgers.edu/cb6le5c1}} As for the original \textbf{SitEnt} dataset, we report class-based F1-scores for \textbf{SitEnt-ambig}.

In order to cover a wider variety of genres, we also evaluate our approach on the \textbf{Captions} dataset of Alikhani and Stone~\shortcite{alikhani-stone-2019-caption}. The dataset is based on a number of image captions corpora and contains annotations for verbs being used as \emph{states}, \emph{telic events} and \emph{atelic events}. For this experiment we merge the \emph{telic} and \emph{atelic} class, resulting in a 2-class problem with 2687 instances with a class distribution of 22:78 (\emph{state}:\emph{event}). The dataset does not contain pre-defined training/evaluation splits. We therefore evaluate using 10-fold cross-validation and report class-based F1-scores.

Finally, we evaluate on our proposed \textbf{DIASPORA} dataset, again merging the \emph{telic} and \emph{atelic} classes for this experiment, resulting in a class distribution of 43:57 (\emph{state}:\emph{event}). We again report class-based F1-scores over 10-fold cross-validation.

\subsubsection{Results}

Table~\ref{tbl:state_vs_event_results} below shows the results on all datasets. We compare classifying the representation of a verb without any context, the verb with local context, and the full sentence, with a majority-class baseline and previous results in the literature. The results of using a local context are based on the best performing context window around the verb, an overview of the effect of the size of the context window is shown in Figure~\ref{fig:state_event_context_window_size}. A result table comparing the best linear context window window with the best performing dependency context window is presented in Table~\ref{tbl:dep_vs_window_state_event} in Appendix~\ref{sec:supplemental_c}.

\begin{table*}[!htb]
\centering
\small
%\resizebox{\textwidth}{!}{
\begin{tabular}{ l | c c c | c c c c}
\textbf{Dataset} 		& \textbf{Verb only}	& \textbf{Local Context}	& \textbf{Full Sentence}		& \textbf{Maj. Class}	& \textbf{FP14}	& \textbf{H16}	& \textbf{F16}\\\hline
Asp-ambig Accuracy & 65.9 & \textbf{74.2} & 60.0 & 65.9 & 72.0 & 72.8 & - \\\hline
SitEnt F1 (State) & \textbf{84.0} & 81.3 & 26.4 & 0.0 & - & - & 80.6\\
SitEnt F1 (Event) & \textbf{86.6} & 84.5 & 71.9 & 68.9 & - & - & 78.6\\\hline
SitEnt-ambig F1 (State) & 44.0 & \textbf{62.6} & 0.0 & 0.0 & - & - & - \\
SitEnt-ambig F1 (Event) & 62.4 & 66.2 & 68.3 & \textbf{68.4} & - & - &.- \\\hline
Captions F1 (State) & 0.1 ($\pm$ 0.07) & \textbf{58.8 ($\pm$ 0.05)} & 23.4 ($\pm$ 0.06) & 0.0 & - & - & - \\
Captions F1 (Event) & 87.3 ($\pm$ 0.01) & \textbf{89.7 ($\pm$ 0.02)} & 86.7 ($\pm$ 0.02) & 87.6 & - & - & - \\\hline
DIASPORA F1 (State) & 76.4 ($\pm$ 0.07) & \textbf{86.5 ($\pm$ 0.04)} & 80.1 ($\pm$ 0.06) & 0.0 & - & - & - \\
DIASPORA F1 (Event) & 83.5 ($\pm$ 0.03) & \textbf{89.8 ($\pm$ 0.03)} & 84.8 ($\pm$ 0.03) & 72.5 & - & - & - \\\hline
\end{tabular}%}
\caption{Results on classifying \emph{states} vs. \emph{events}. \textbf{FP14} refers to Friedrich and Palmer~\shortcite{Friedrich_2014}, \textbf{H16} to Heuschkel~\shortcite{Heuschkel_2016}, and \textbf{F16} to Friedrich et al.~\shortcite{Friedrich_2016}.}
\label{tbl:state_vs_event_results}
\end{table*}

In general, a local context window exhibits the strongest performance, even achieving a new state-of-the-art on the \textbf{Asp-ambig} dataset, despite the simplicity of our setup. The strong results of the verb-only model on the \textbf{SitEnt} dataset, that substantially outperforms the sequence model of Friedrich et al.~\shortcite{Friedrich_2016} and the local context model, confirms our suspicion that the classifier learnt the fact that most verbs in the dataset occur unambiguously with their target label. This is furthermore reflected in the results on the \textbf{SitEnt-ambig} dataset, where using only the verb leads to considerably worse performance than when taking a local context window around the verb into account. While the results on \textbf{SitEnt-ambig} are generally low, this reflects the increased difficulty of the task as well as the simplicity of our setup, and we expect to improve on these results with higher capacity models in future work.

\subsubsection{Analysis}

\begin{figure*}[!htb]
\centering
\includegraphics[width=\textwidth]{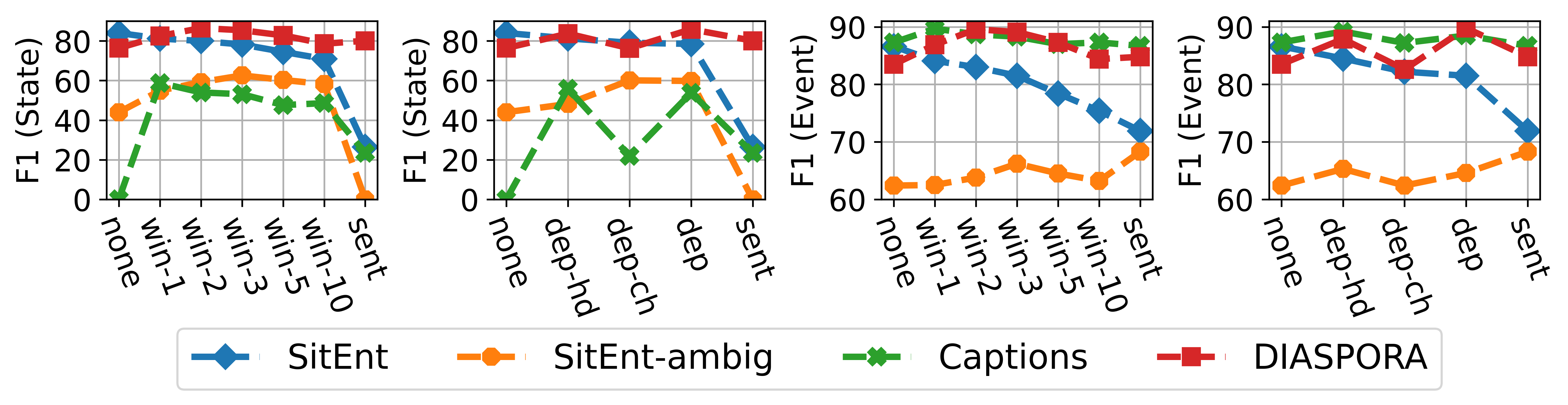}
\captionsetup{font=small}
\caption{Class-based F1-score performance trajectories for varying sizes of the context window across all datasets.}
\label{fig:state_event_context_window_size}
\end{figure*}

In Figure~\ref{fig:state_event_context_window_size} we show class-based F1-score performance trajectories for varying sizes of the linear context window and the dependency context across all datasets. We observe that performance typically peaks at a narrow context window of taking 1-3 surrounding words into account, with performance dropping steeply when increasing the context window.\footnote{We note that the F1-score for event on the \textbf{SitEnt-ambig} datasets appears to peak by taking the whole sentence as input, however this is an effect of its performance degrading to that of the majority class baseline, as can also be seen on its near-0 performance for predicting states.} Our results also exhibit that linear window contexts are typically better predictors for predicational aspect than dependency contexts.

\begin{figure*}[!htb]
\centering
\includegraphics[width=\textwidth]{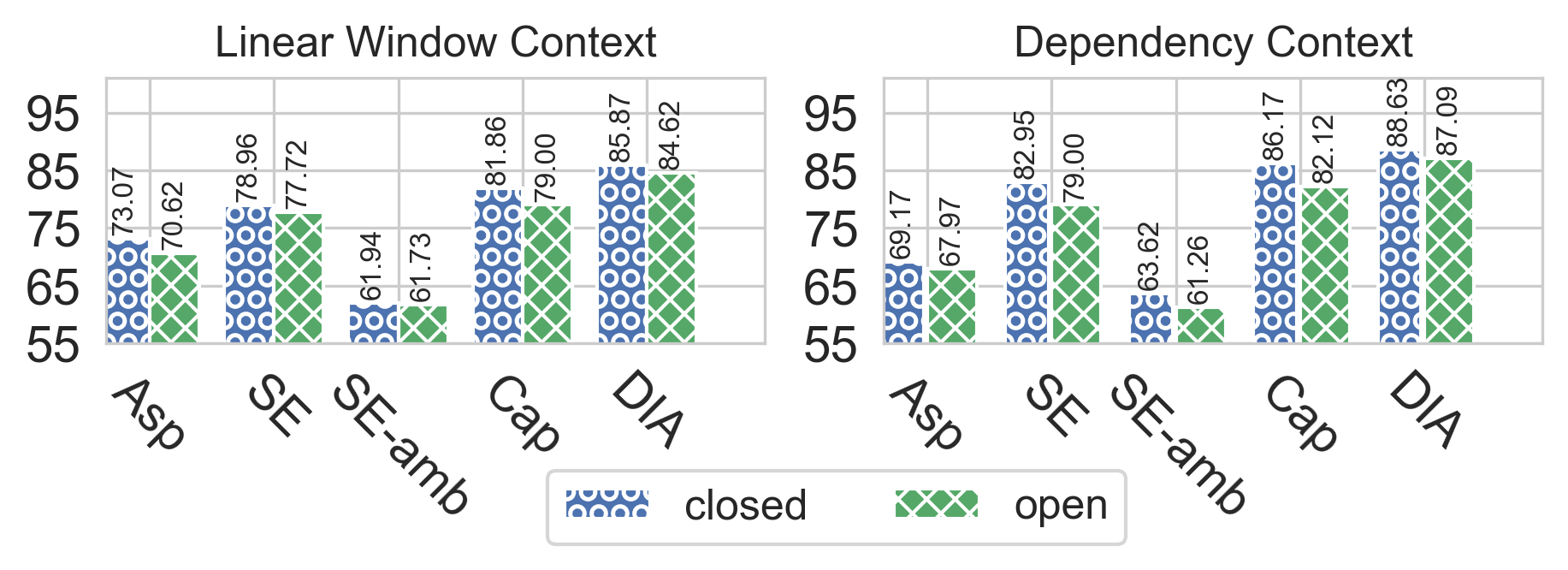}
\captionsetup{font=small}
\caption{Averaged accuracy scores for closed class context words in comparison to content word contexts.}
\label{fig:pos_tag_acc_state_event}
\end{figure*}

This is an interesting result as dependency contexts are more likely to yield content words, such as nouns, adjectives or other verbs as context,\footnote{As one of our reviewers pointed out, this likely is due to our use of Universal Dependencies~\cite{Osborne_2019}. We show the distribution of extracted context words for the linear and dependency context windows per PoS tag in Appendix~\ref{sec:supplemental_b}.} as opposed to linear context windows yielding more closed class words. We investigate this effect further by comparing the general overall performance of closed class words with content words.
Figure~\ref{fig:pos_tag_acc_state_event} provides empirical evidence that closed class words are strong predictors of predicational aspect. The figure shows accuracies for PoS tags belonging to a closed-class group, in comparison to ones belonging to open class content words. We calculated PoS-based accuracy by counting how often a word with a given PoS tag contributed to a correct classification as opposed to an incorrect one. For example if the PoS tag \texttt{IN}\footnote{\texttt{IN} refers to a preposition or subordinating conjunction.} occurs 8 times as part of correctly classified context windows and 2 times as part of incorrectly classified ones, we estimate its accuracy as 0.8. We count the participation of a PoS tag for a correct or incorrect classification decision as evidence that the given word is a reliable predictor for a given class. We expect that words with high predictive capacity will more often occur in correctly classified context windows than in incorrectly classified ones.

Figure~\ref{fig:pos_tag_acc_state_event} highlights that closed class words are typically more reliable predictors for predicational aspect than content words. This is a very interesting result, given our model solely operates on the basis of composed word vectors, thus indicating that distributional representations for closed class words encode a substantial amount of information that can potentially be leveraged for fine-grained directional inferences. In order to assess the generalisation capability of distributional representations we performed a zero-shot experiment on the \textbf{Asp-ambig} dataset where we held out all annotated data for a specific verb for evaluation, and trained the model on the remaining data. Table~\ref{tbl:asp_ambig_zero_shot} in Appendix~\ref{sec:supplemental_d} provides evidence that distributional representations capture predicational aspect of unseen verbs to a surprising extent.

\begin{table*}[!htb]
\centering
\begin{tabular}{ c | c | l }
\textbf{Aspect} & \textbf{Verb}	& \textbf{Example Sentences}\\\hline
Event & look & (1) Shane \emph{looked at} his watch.\\
State & look & (2) She sure \emph{looks} good.\\
Event & stand & (3) Jefferson shook his head and \emph{stood up}.\\
State & stand & (4) A metal table and four chairs \emph{stood in} the center.\\
\end{tabular}%}
\caption{Example Sentences from the state vs. event datasets.}
\label{tbl:example_sentences_state_event}
\end{table*}

Table~\ref{tbl:example_sentences_state_event} shows example sentences for two ambiguous verbs from our datasets. In the first and third sentence the preposition \emph{at} and the particle \emph{up}, respectively, cause the predicate to express an event. Without a preposition, verbs such as \emph{look} can express stative aspect as in the second sentence. The last sentence is an interesting case where the verb \emph{stand} occurs in the context of a preposition, yet the combination remains stative, as the sentence describes the arrangement of inanimate objects.

\subsection{Experiment 2 --- Telic vs. Atelic Events}
\label{sec:exp_telic_vs_atelic}

For classifying \emph{telic} and \emph{atelic} events we are using the \emph{Telicity} dataset of Friedrich and Gateva~\shortcite{Friedrich_2017}, the \textbf{Captions} dataset of Alikhani and Stone~\shortcite{alikhani-stone-2019-caption}, as well as our own proposed \textbf{DIASPORA} dataset.

The \textbf{Telicity} dataset contains 1863 sentences extracted from the MASC corpus, where a verb in context is labelled as either \emph{telic} or \emph{atelic}. The dataset is imblanced with 82\% of verb occurrences being labelled as \emph{telic}. We follow the experimental protocol of Friedrich and Gateva~\shortcite{Friedrich_2017} and report accuracy and class-based F1-scores, using document-based cross-validation. During our experimental work we again noticed that only 70 out of approximately 570 distinct verbs in the dataset occur with both labels. However, applying the same strategy as for the \textbf{SitEnt-ambig} dataset would have resulted in too little data.\footnote{Using only sentences with verbs that occur with both labels regardless of their class distribution resulted in only 295 examples in total.} Therefore, given this characteristic, we again expect the classifier using the verb without any context to achieve artificially high performance.

For the \textbf{Captions} dataset, we omit the examples labelled as \emph{stative}, leaving us with 2092 captions in total, of which 800 are annotated as \emph{telic} (38\%), and 1292 as \emph{atelic} (62\%). We perform 10-fold cross-validation and report accuracy and class-based F1-scores.

Finally, for our \textbf{DIASPORA} dataset, we also omit the utterances annotated as expressing \emph{stative} predicational aspect, leaving us 527 examples in total, 279 instances labelled as \emph{telic} (53\%), and 248 instances labelled as \emph{atelic} (47\%), thus representing the most balanced dataset among the three.

\subsubsection{Results}

Table~\ref{tbl:telic_vs_atelic_events} shows the results for all three datasets, comparing a model that only has access to the distributional representation of the target verb itself, with models that have access to a local context window and the full sentence, as well as to previous results in the literature. A result table comparing the best linear context window window with the best performing dependency context window is presented in Table~\ref{tbl:dep_vs_window_telic_atelic} in Appendix~\ref{sec:supplemental_c}.

\begin{table*}[!htb]

\centering
\small
%\resizebox{\textwidth}{!}{
\begin{tabular}{ l | c c c | c c c}
\textbf{Dataset} 		& \textbf{Verb only}	& \textbf{Local Context}	& \textbf{Full Sentence}		& \textbf{Maj. Class}	& \textbf{FG17} & \textbf{FG17+IC}	\\\hline
Telicity Accuracy   & \textbf{87.2} & 86.4 & 85.8 & 82.0 & 86.7 & 82.3  \\
Telicity F1 (Telic) & 92.1 & 91.6 & 91.1 & 90.1 & \textbf{92.2} & 88.6  \\
Telicity F1 (Atelic)& \textbf{62.6} & 60.3 & 60.6 & 0.0  & 53.7 & 61.4 \\\hline
Captions Accuracy   & 78.8 ($\pm$ 0.03) & \textbf{79.1 ($\pm$ 0.02)} & 78.8 ($\pm$ 0.02) & 61.8 & - & -  \\
Captions F1 (Telic) & 70.7 ($\pm$ 0.05) & \textbf{72.3 ($\pm$ 0.03)} & 71.2 ($\pm$ 0.04) & 55.3 & - & -  \\
Captions F1 (Atelic)& \textbf{83.3 ($\pm$ 0.03)} & 83.0 ($\pm$ 0.02) & 83.1 ($\pm$ 0.02) & 0.0 & - & -  \\\hline
Dialogue Accuracy   & 64.3 ($\pm$ 0.04) & \textbf{69.3 ($\pm$ 0.05)} & 65.1 ($\pm$ 0.05) & 52.9 & - & -  \\
DIASPORA F1 (Telic) & 64.5 ($\pm$ 0.06) & \textbf{70.3 ($\pm$ 0.06)} & 66.5 ($\pm$ 0.05) & 69.2 & - & - \\
DIASPORA F1 (Atelic)& 63.3 ($\pm$ 0.05) & \textbf{67.8 ($\pm$ 0.05)} & 62.9 ($\pm$ 0.07) & 0.0 & - & - \\\hline
\end{tabular}%}
\caption{Results on classifying telic vs. atelic events. \textbf{FG17} refers to the best performing model of Friedrich and Gateva~\shortcite{Friedrich_2017}, and \textbf{FG17+IC} refers to the model of Friedrich and Gateva~\shortcite{Friedrich_2017} with access to additional data.}
\label{tbl:telic_vs_atelic_events}
\end{table*}

Our purely distributional models achieve competitive results, with the expected strong performance for the verb-only model, that is even beating the current state-of-the-art in terms of accuracy and F1-score for the \emph{atelic} class. For the \textbf{Captions} and \textbf{DIASPORA} datasets we observe similar trends as for the \emph{state} vs. \emph{event} datasets above, with the models that operate over a local context window typically achieving the strongest performance. Notably, the verb-only models are able to perform competitively with local context windows across all datasets. While telicity itself is not part of the morphology of English verbs, telic events frequently correlate with the past tense, such that the distributional representation for the inflected verb already encodes a substantial amount of information. 

\subsubsection{Analysis}

\begin{figure*}[!htb]
\centering
\includegraphics[width=\textwidth]{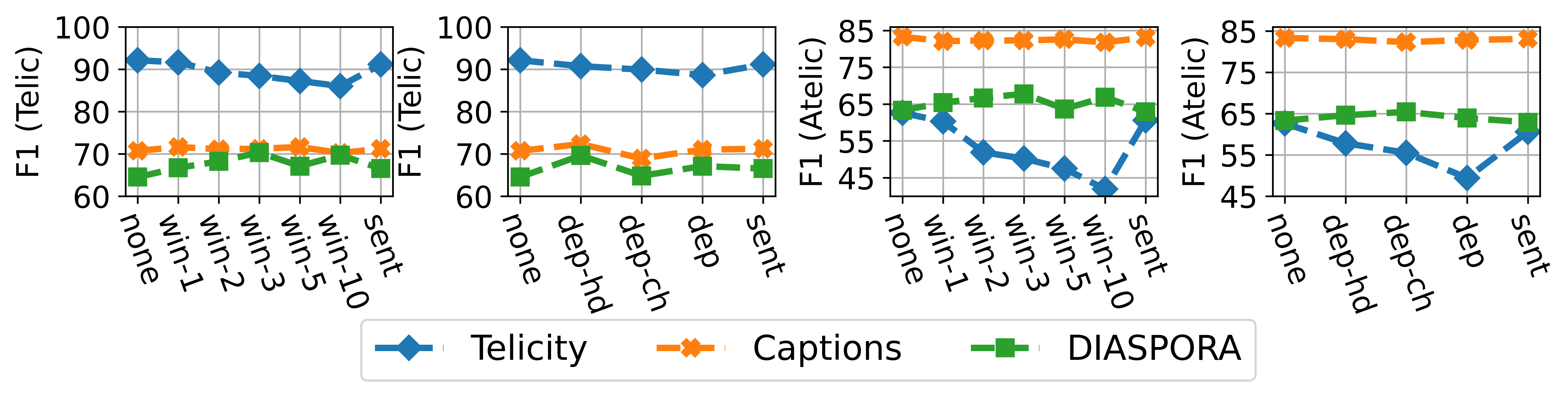}
\captionsetup{font=small}
\caption{Class-based F1-score performance trajectories for varying sizes of the context window across all datasets.}
\label{fig:telic_atelic_context_window_size}
\end{figure*}

Figure~\ref{fig:telic_atelic_context_window_size} shows a class-based F1-score performance trajectory across all datasets and varying context window sizes. Unlike for distinguishing states from events in Figure~\ref{fig:state_event_context_window_size} above, predicting telicity appears to be less dependent on a small local context window surrounding the target verb. This is reflected in Figure~\ref{fig:telic_atelic_context_window_size} which does not contain such clear performance peaks, but is more uniform across different sizes of context windows.

\begin{figure*}[!htb]
\centering
\includegraphics[width=\textwidth]{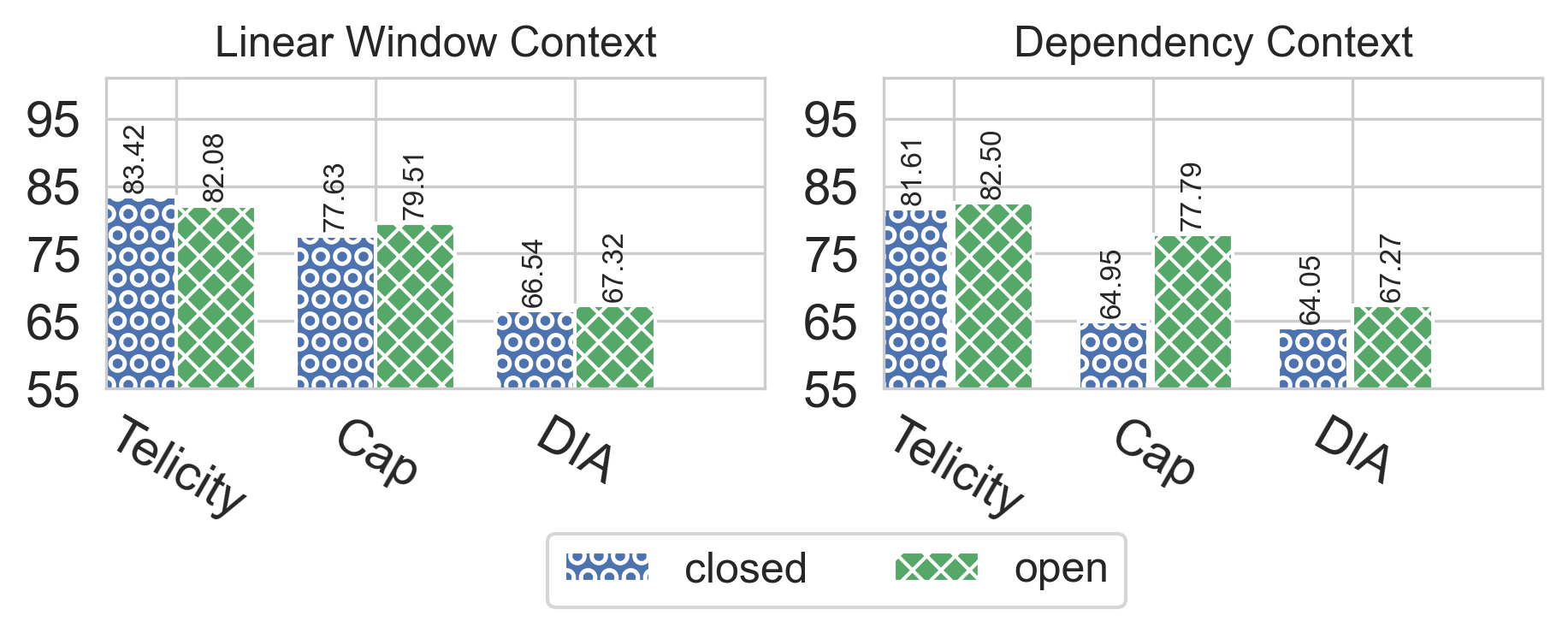}
\captionsetup{font=small}
\caption{Averaged accuracy scores for closed class context words in comparison to content word contexts.}
\label{fig:pos_tag_acc_telic_atelic}
\end{figure*}

We furthermore show the averaged PoS-based accuracy plot in Figure~\ref{fig:pos_tag_acc_telic_atelic}. For predicting telicity, closed class words are less reliable predictors in comparison to content words than for modelling states and events above. 

This result becomes more transparent when analysing actual sentences from our dataset.

\begin{table*}[!htb]
\centering
%\resizebox{\textwidth}{!}{
\begin{tabular}{ c | c | l }
\textbf{Aspect} & \textbf{Verb}	& \textbf{Example Sentences}\\\hline
Telic & leave & (1) Okay, I \emph{have left} the building.\\
Atelic & walk & (2) Okay, I\emph{'m still walking} towar- oh is it blue? \\
Telic & turn & (3) Fans \emph{turned} on the players and manager.\\
Atelic & paddle & (4) Four kayakers \emph{paddle} through the water.\\
\end{tabular}%}
\caption{Example Sentences from the telic vs. atelic datasets.}
\label{tbl:example_sentences_telic_atelic}
\end{table*}

Table~\ref{tbl:example_sentences_telic_atelic} shows some example sentences from the datasets annotated for telicity. The sentences show that telicity in English is frequently associated with tense, with present tenses indicating atelic eventualities and past tense indicating a completed event. This suggests that frequently the verb by itself might be sufficient for inferring telicity as in sentences (3) and (4). In many other cases, the verb interacts with its auxiliary in a tensed construction as in sentences (1) and (2).

\section{Conclusion}
\label{conclusion}

In this work, we have proposed the first dataset of human-human dialogues annotated with the aspectual class of verbs. 
We have proposed a compositional distributional approach for modelling the aspectual class of English verbs in context. Our results indicate that distributional models are able to learn concise representations for closed class words such as particles and prepositions, and that classifiers using composed distributional representations achieve a new state-of-the-art on three recently proposed datasets. We have furthermore contributed a qualitative analysis, providing empirical evidence for the long standing insight of semanticists that the presence of prepositions or particles in a verb phrase, tend to be very reliable indicators of the verb's aspectual class~(\newcite{Vendler_1957},~\newcite{Dowty_1979},~\newcite{Moens_1988}, \emph{passim}). Our model setup was intentionally kept simple as we were primarily concerned with the question whether predicational aspect can be captured with a distributional semantics approach \emph{in principle}. We note that using more sophisticated models might yield even stronger results, although in preliminary tests, we did not observe any meaningful performance difference when replacing our {bag-of-embeddings} approach with ELMo~\cite{Peters_2018} or BERT~\cite{Devlin_2019}.

While this work was done on English, we aim to use our methodology in a multilingual setup in future work as distributional approaches scale well with growing amounts of data and across languages.% For example the Slavic language family is an interesting test bed for future work due to explicit morphological markers for telicity.

Aspect, alongside tense, is a crucial indicator of the temporal extent of a verb as well as the entailments it licenses. In future work we plan to integrate aspectual information for improving the unsupervised construction of entailment graphs~\cite{Berant_2010,Hosseini_2018}, as well as temporal reasoning, which has been shown recently to be difficult for distributional semantic models~\cite{Kober_2019}.

Aspectual information can be utilised for directional entailment detection by inferring that the event of \emph{buying something} entails the state of \emph{owning that thing}, but not the other way round. Determining the \emph{telicity} of an event also enables fine-grained inferences about whether an event caused a change of state. For example, while the telic context of \emph{writing a sonnet in fifteen minutes} entails a change to a state where a finished sonnet exists, the atelic context of \emph{writing a sonnet for fifteen minutes} does not.
\section*{Acknowledgements}
\label{acknowledgements}

The research presented here is supported by NSF IIS-1526723 and CCF-19349243 and in part by ERC Advanced Fellowship 742137 SEMANTAX.

\bibliographystyle{coling}

\bibliography{coling2020}

\clearpage
\appendix

\section{Supplemental Material --- Performance per Verb Type in Asp-ambig}
\label{sec:supplemental_a}
Table~\ref{tbl:asp_ambig_results_per_verb} below lists the results per verb type for all 20 ambiguous verbs in the \textbf{Asp-ambig} dataset, comparing the majority class baseline, the models of Friedrich and Palmer~\shortcite{Friedrich_2014} and Heuschkel~\shortcite{Heuschkel_2016} to our window-1 and dependency (full) approaches.

\begin{table}[!htb]
\centering
\small
\resizebox{\textwidth}{!}{
\begin{tabular}{ l | c | c | c | c | c}
\textbf{Verb} & \textbf{Majority class} & \textbf{Friedrich and Palmer~\shortcite{Friedrich_2014}} & \textbf{Heuschkel~\shortcite{Heuschkel_2016}} & \textbf{window-1} & \textbf{dependency (full)}\\ \hline
\emph{feel} 	& \textbf{96.1} & 93.8 & 95.5 & 94.2 & 93.5 \\
\emph{say} 		& \textbf{94.9} & 93.5 & 94.2 & \textbf{94.9} & \textbf{94.9}\\
\emph{make} 	& 91.9 & 91.2 & \textbf{92.0} & 90.6 & 90.6 \\
\emph{come} 	& \textbf{88.0} & 87.2 & \textbf{88.0} & 85.5 & 86.2\\
\emph{take} 	& 85.4 & 85.4 & 85.5 & 85.5 & \textbf{87.0}\\
\emph{meet} 	& 83.9 & 87.7 & 85.9 & 87.7 & \textbf{90.6}\\
\emph{stand} 	& 80.0 & 83.1 & 81.8 & \textbf{87.7} & 82.6\\
\emph{find} 	& 74.5 & 68.8 & \textbf{76.6} & 75.4 & 69.6\\
\emph{accept} 	& \textbf{70.9} & 65.7 & 67.4 & 68.8 & 61.6\\
\emph{hold} 	& 56.0 & 49.3 & 57.8 & \textbf{62.3} & 55.8 \\
\emph{carry} 	& 55.9 & 58.1 & 60.3 & 58.0 & \textbf{63.8}\\
\emph{look} 	& 55.8 & 74.6 & 65.9 & \textbf{79.7} & 61.6\\
\emph{show} 	& 54.9 & \textbf{68.4} & 65.9 & 67.4 & 65.2\\
\emph{appear} 	& 52.2 & 61.0 & 56.6 & \textbf{70.3} & 62.3\\
\emph{follow} 	& 51.6 & 65.6 & 61.8 & 61.6 & \textbf{69.6} \\
\emph{consider} & 50.7 & 70.3 & 67.4 & \textbf{77.5} & 59.4\\
\emph{cover} 	& 50.4 & 54.5 & 55.0 & \textbf{57.2} & 50.0\\
\emph{fill} 	& 47.8 & 62.7 & \textbf{69.4} & 60.1 & 48.6\\
\emph{bear} 	& 47.4 & 67.4 & \textbf{77.2} & 68.1 & 63.8\\
\emph{allow} 	& 37.8 & \textbf{51.9} & 51.1 & 50.7 & 44.9\\\hline
Average			& 66.3 & 72.0 & 72.8 & \textbf{74.2} & 70.1 \\
\end{tabular}}
\captionsetup{font=small}
\caption{Per verb Accuracies on the \textbf{Asp-ambig} dataset~\cite{Friedrich_2014}.}
\label{tbl:asp_ambig_results_per_verb}
\end{table}

For strongly imbalanced classes as in the case of \emph{feel}, which almost always functions as a \emph{state}, the majority baseline is very difficult to beat. Interestingly, the window-1 and dependency (full) approaches frequently exhibit complementary performance. For example, while for \emph{stand} or \emph{look} a window-based context works substantially better, for \emph{follow} or \emph{carry} a dependency-based context is preferable. One explanation for this behaviour is that for \emph{stand} or \emph{look} prepositions are frequently the most salient indicator of aspectual class as shown in Section~\ref{sec:experiments}. On the other hand, for \emph{follow} or \emph{carry} a content word, such as the subject or direct object, is frequently more salient.
\clearpage
\section{Supplemental Material --- PoS Tag Distribution of Extracted Contexts}
\label{sec:supplemental_b}
\begin{figure*}[!htb]
\centering
\includegraphics[width=\textwidth]{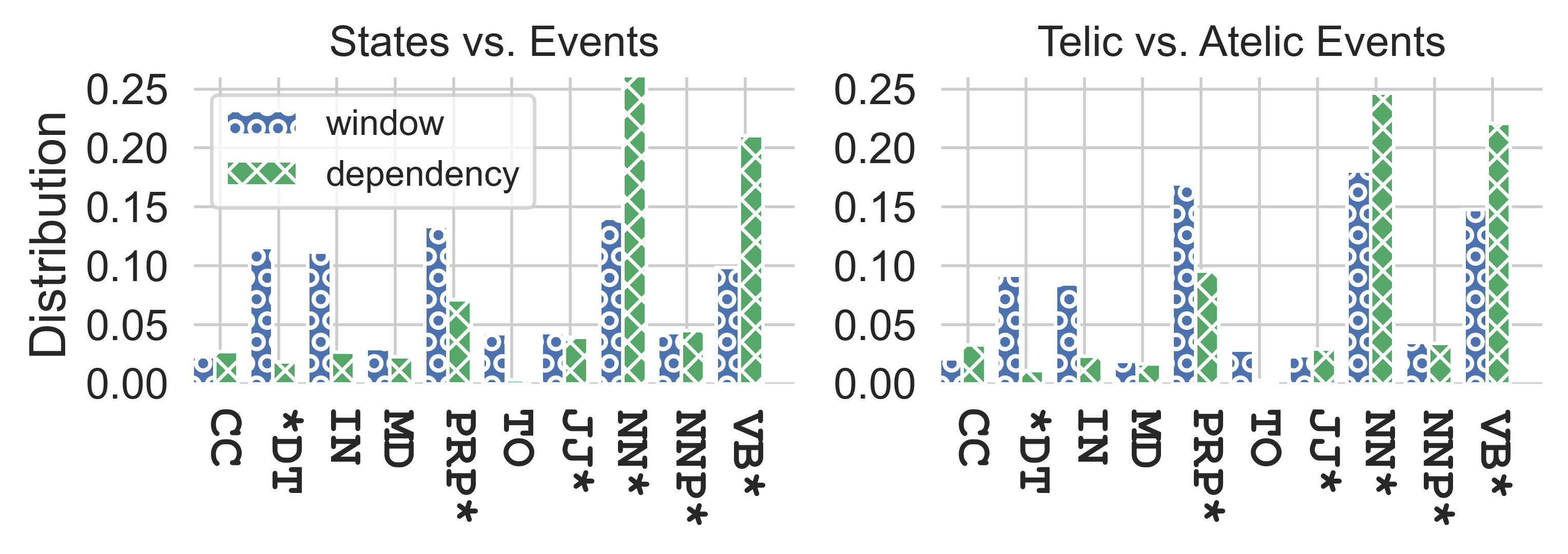}
\captionsetup{font=small}
\caption{PoS tag distribution of the extracted contexts of the linear context window vs. dependency contexts.}
\label{fig:pos_distribution_window_vs_dependency}
\end{figure*}

Figure~\ref{fig:pos_distribution_window_vs_dependency} shows the PoS tag distribution of extracted contexts of the linear context window in comparison to dependency contexts. Dependency contexts, based on Universal Dependencies, overwhelmingly extract content words, whereas the linear context window predominantly tends to extract more closed class words.
\clearpage
\section{Supplemental Material --- Window Contexts vs. Dependency Contexts}
\label{sec:supplemental_c}
Tables~\ref{tbl:dep_vs_window_state_event} \&~\ref{tbl:dep_vs_window_telic_atelic} present the performance of the best performing linear context window in comparison to the best performing dependency context window, as well as the verb-only and full-sentence and majority class baselines.

% State vs Event
\begin{table*}[!htb]
\centering
\small
%\resizebox{\textwidth}{!}{
\begin{tabular}{ l | c c | c c c}
\textbf{Dataset} 		& \textbf{Best Linear}	& \textbf{Best Dependency}	& \textbf{Verb only}		& \textbf{Full Sentence}	& \textbf{Majority Class}\\\hline
Asp-ambig Accuracy & \textbf{74.2} & 70.1 & 65.9 & 60.0 & 65.9\\\hline
SitEnt F1 (State) & 81.2 & 81.3 & \textbf{84.0} & 26.4 & 0.0\\
SitEnt F1 (Event) & 84.1 & 84.5 & \textbf{86.6} & 71.9 & 68.9\\\hline
SitEnt-ambig F1 (State) & \textbf{62.6} & 60.1 & 44.0 & 0.0 & 0.0\\
SitEnt-ambig F1 (Event) & 66.2 & 65.3 & 62.4 & 68.3 & \textbf{68.4} \\\hline
Captions F1 (State) & \textbf{58.8} & 55.9 & 0.1 & 23.4 & 0.0 \\
Captions F1 (Event) & \textbf{89.7} & 89.2 & 87.3 & 86.7 & 87.6\\\hline
DIASPORA F1 (State) & \textbf{86.5} & 85.9 & 76.4 & 80.1 & 0.0 \\
DIASPORA F1 (Event) & 89.6 & \textbf{89.8} & 83.5 & 84.8 & 72.5\\\hline
\end{tabular}%}
\caption{Comparison between linear contexts windows and dependency contexts on classifying \emph{states} vs. \emph{events}.}
\label{tbl:dep_vs_window_state_event}
\end{table*}
Overall linear context window perform slightly better on average than dependency context windows --- and as highlighted in Section~\ref{sec:experiments} --- this can be explained by linear context window extracting more closed class context words (see Figure~\ref{fig:pos_distribution_window_vs_dependency} in Appendix~\ref{sec:supplemental_b}), which tend to be stronger disambiguation signals than content words.
% Telic vs Atelic
\begin{table*}[!htb]

\centering
\small
%\resizebox{\textwidth}{!}{
\begin{tabular}{ l | c c | c c c}
\textbf{Dataset} 		& \textbf{Best Linear}	& \textbf{Best Dependency}	& \textbf{Verb only}		& \textbf{Full Sentence}	& \textbf{Majority Class}\\\hline
Telicity F1 (Telic) & 91.6 & 90.7 & \textbf{92.1} & 91.1 & 90.1  \\
Telicity F1 (Atelic)& 60.3 & 57.8 & \textbf{62.6} & 60.6  & 0.0 \\\hline
Captions F1 (Telic) & 71.6 & \textbf{72.3} & 70.7 & 71.2 & 55.3 \\
Captions F1 (Atelic)& 82.6 & 83.0 & \textbf{83.3} & 83.1 & 0.0 \\\hline
DIASPORA F1 (Telic) & \textbf{70.3} & 69.6 & 64.5 & 66.5 & 69.2 \\
DIASPORA F1 (Atelic)& \textbf{67.8} & 65.4 & 63.3 & 62.9 & 0.0 \\\hline
\end{tabular}%}
\caption{Comparison between linear contexts windows and dependency contexts on classifying \emph{telic} vs. \emph{atelic} events.}
\label{tbl:dep_vs_window_telic_atelic}
\end{table*}
\clearpage
\section{Supplemental Material --- Zero Shot Generalization}
\label{sec:supplemental_d}
For assessing the generalisation capabilities of our methodology we are performing a zero-shot setup on the \textbf{Asp-ambig} dataset. Instead of running an evaluation for each verb individually as originally proposed by Friedrich and Palmer~\shortcite{Friedrich_2014}, we are evaluting the model on the data for one particular verb, say \emph{look}, and train the model on all available data, \textbf{except} the data for the heldout verb \emph{look}. This way, we investigate whether distributional representations truly capture the underlying semantics of predicational aspect.

We use the same simple setup as in Section~\ref{sec:experiments}, with a linear regression classifier that operates on the basis of averaged word2vec embeddings. We used a linear context window of size 1 for this experiments as this was the best performing setup for the \textbf{Asp-ambig} dataset in the evaluation in Section~\ref{sec:experiments}.
\begin{table}[!htb]
\centering
\small
%\resizebox{\textwidth}{!}{
\begin{tabular}{ l | c | c }
\textbf{Verb} & \textbf{Majority class} & \textbf{Zero Shot} \\ \hline
\emph{feel} 	& \textbf{96.1} & 43.0 \\
\emph{say} 		& \textbf{94.9} & 55.8 \\
\emph{make} 	& \textbf{91.9} & 86.2 \\
\emph{come} 	& \textbf{88.0} & 78.1 \\
\emph{take} 	& \textbf{85.4} & 82.6\\
\emph{meet} 	& \textbf{83.9} & 76.8 \\
\emph{stand} 	& \textbf{80.0} & 22.5\\
\emph{find} 	& \textbf{74.5} & 69.6\\
\emph{accept} 	& \textbf{70.9} & 68.1\\
\emph{hold} 	& \textbf{56.0} & 26.3 \\
\emph{carry} 	& \textbf{55.9} & 54.0 \\
\emph{look} 	& 55.8 & \textbf{77.5} \\
\emph{show} 	& \textbf{54.9} & 52.2\\
\emph{appear} 	& \textbf{52.2} & 51.4\\
\emph{follow} 	& \textbf{51.6} & 39.9 \\
\emph{consider} & \textbf{50.7} & \textbf{50.7}\\
\emph{cover} 	& \textbf{50.4} & 41.6\\
\emph{fill} 	& 47.8 & \textbf{48.2}\\
\emph{bear} 	& \textbf{47.4} & 47.1\\
\emph{allow} 	& 37.8 & \textbf{39.4}\\\hline
Average			& \textbf{66.3} & 55.5 \\
\end{tabular}%}
\captionsetup{font=small}
\caption{Per verb Accuracies on the \textbf{Asp-ambig} dataset~\cite{Friedrich_2014}.}
\label{tbl:asp_ambig_zero_shot}
\end{table}
Table~\ref{tbl:asp_ambig_zero_shot} shows the results of the zero-shot experiment in comparison to the majority class baseline. While for the majority of verbs, our model underperforms the majority class baseline --- which is difficult to beat especially for the very skewed verbs such as \emph{feel} or \emph{say}, our approach beats the baseline for 3 verbs and achieves comparable performance for more than half of the verbs, while not having encountered any annotated data for the target verb during training at all. 

Given the simplicity of our setup, we regard that as strong evidence that a model based on distributional semantics does indeed capture a substantial amount of predicational aspect in its representations. 

\end{document}